# Morphological Computation and Learning to Learn In Natural Intelligent Systems And AI

Gordana Dodig-Crnkovic [1,2]

**Abstract.** At present, artificial intelligence in the form of machine learning is making impressive progress, especially the field of deep learning (DL) [1]. Deep learning algorithms have been inspired from the beginning by nature, specifically by the human brain, in spite of our incomplete knowledge about its brain function. Learning from nature is a two-way process as discussed in [2][3][4], computing is learning from neuroscience, while neuroscience is quickly adopting information processing models. The question is, what can the inspiration from computational nature at this stage of the development contribute to deep learning and how much models and experiments in machine learning can motivate, justify and lead research in neuroscience and cognitive science and to practical applications of artificial intelligence.

## 1 INTRODUCTION

This paper explores the relationships between the info-computational network based on morphological computation and the present developments in both the sciences of the artificial (with the focus on deep learning) as well as natural sciences (especially neuroscience, cognitive science and biology), social sciences (social cognition) and philosophy (philosophy of computing and philosophy of mind).

Deep learning is based on artificial neural networks resembling neural networks of the brain, processing huge amounts of (labelled) data by high-performance GPUs (graphical processing units) with a parallel architecture. It is (typically supervised) machine learning from examples. It is static, based on the assumption that the world behaves in a similar way and that domain of application is close to the training data. However impressive and successful, deep-learning intelligence has an Achilles heel, and that is lack of common sense reasoning [5][6][7]. It bases recognition of pictures on pixels, and small changes, even invisible for humans can confuse deep learning algorithm and lead to very surprising errors.

According to Bengio, deep learning is missing out of distribution generalization, and compositionality. Human intelligence has two distinct mechanisms of learning – quick, bottom up, from data to patterns (System 1) and slow, top-down from language to objects (System 2) which have been recognized earlier [8][9][10]. The starting point of old AI (GOFAI) was System 2, symbolic, language, logic-based reasoning, planning and decision making. However, it was without System 1 so it ended in symbol grounding problem. Now deep learning has grounding for its symbols in the data, but it lacks the System 2 capabilities in order to get to the human-level intelligence and ability of learn and meta-learning, that is learning to learn.

The step from big-data based System 1 to manipulation of few concepts like in high level reasoning is suggested to proceed via concepts of agency, attention and causality.

It is expected that agent perspective will help to put constraints on the learned representations and so to encapsulate causal variables, and affordances. Bengio proposes that "meta-learning, the modularization aspect of the consciousness prior [7] and the agent perspective on representation learning should facilitate re-use of learned components in novel ways (even if statistically improbable, as in counterfactuals), enabling more powerful forms of compositional generalization, i.e., out-of-distribution generalization based on the hypothesis of localized (in time, space, and concept space) changes in the environment due to interventions of agents." [5]

This step, from System 1 (present) to System 2 (higher level cognition will open new and even more powerful possibilities to AI. It is not the development into the unknown, as some of it has been attempted by GOFAI, and new developments in cognitive science and neuroscience. In this article we will focus on the connections to another computational model of cognition, natural infocomputation [3][4].

## 1 LEARNING ABOUT THE WORLD THROUGH AGENCY

When discussing cognition as a bioinformatic process of special interest, we use the notion of *agent*, i.e. a *system able to act on its own behalf* [11]. Agency in biological systems in the sense I use here has been explored in [12][13]. The world as it appears to an agent depends on the type of interaction through which the agent acquires information [11].

Agents communicate by exchanging messages (information) which helps them to coordinate their actions based on the information they possess and then they share through social cognition.

We start from the definition of *agency and cognition as a property of all living organisms,* building on Maturana and Varela [13][14] and Stewart [16]. The next question will be how artifactual agents should be built in order to possess cognition and eventually even consciousness. Is it possible at all, given that cognition in living organisms is a deeply biologically rooted process? Along with reasoning, language is considered high-level cognitive activity that only humans are capable of. Increasing levels of cognition evolutionary developed in living organisms, starting from basic automatic behaviours such as found in bacteria to insects (even though they have nervous system and brain, they lack the limbic system that (in amniota = limbed vertebrates = reptiles, birds and mammals) controls emotional response to physical stimuli, suggesting they don't process physical stimuli emotionally) to increasingly complex behaviour in higher organisms such as mammals. Can AI "jump over" evolutionary steps in the development of cognition?

[1] Department of Computer Science and Engineering at Chalmers University of Technology and the University of Gothenburg, Sweden. Email: `dodig@chalmers.se`

[2] School of Innovation, Design and Engineering, Mälardalen University, Sweden. Email: `gordana.dodig-crnkovic@mdh.se`

The framework for the discussion is the *computing nature* in the form of *info-computationalism*. It takes *the world (Umwelt)* for an agent to be *information* with its *dynamics* seen as *computation*. Information is observer relative and so is computation. [11][17][18]

Cognition has been studied as information processing in such simple organisms as bacteria [19], [20] as well as cognitive processes in other, more complex multicellular life forms. While the idea that *cognition is a biological process in all living organisms* has been extensively discussed [14][16][21], it is not clear on which basis cognitive processes in all kinds of organisms would be accompanied by (some kind of, some degree of) consciousness. If we in parallel with "minimal cognition" [22] search for "minimal consciousness", what would that be? Opinions are divided at what point in the evolution one can say that consciousness emerged. Some would suggest as Liljenström and Århem that only humans possess consciousness, while the others are ready to recognize consciousness in animals with emotions (like amniota) [23][24]. From the info-computational point of view it has been argued that cognitive agents with nervous systems are the step in evolution which first enabled consciousness in the sense of internal model with the ability of distinguishing the "self" from the "other" [4][25].

## 2 LEARNING IN THE COMPUTING NATURE

For naturalist, nature is the only reality [26]. Nature is described through its structures, processes and relationships, using a scientific approach [27][28]. Naturalism studies the evolution of the entire natural world, including the life and development of human and humanity as a part of nature. Social and cultural phenomena are studied through their physical manifestations. An example of contemporary naturalist approach is the research field is social cognition with its network-based studies of social behaviors.

Computational naturalism (pancomputationalism, naturalist computationalism, computing nature)[29][30][31][3][4] is the view that the entire nature is a huge network of computational processes, which, according to physical laws, computes (dynamically develops) its own next state from the current one. Among prominent. representatives of this approach are Zuse, Fredkin, Wolfram, Chaitin and Lloyd, who proposed different varieties of computational naturalism. According to the idea of computing nature, one can view the time development (dynamics) of physical states as information processing (natural computation). Such processes include self-assembly, self-organization, developmental processes, gene regulation networks, gene assembly, protein-protein interaction networks, biological transport networks, social computing, evolution and similar processes of morphogenesis (creation of form). The idea of computing nature and the relationships between two basic concepts of information and computation are explored in [11][17][18].

In the computing nature, cognition is a natural process, seen as a result of natural bio-chemical processes. All living organisms possess some degree of cognition and for the simplest ones like bacteria cognition consists in metabolism and (my addition) locomotion. [11] This "degree" is not meant as continuous function but as a qualitative characterisation that cognitive capacities increase from simplest to the most complex organisms. The process of interaction with the environment causes changes in the informational structures that correspond to the body of an agent and its control mechanisms, which define its future interactions with the world and its inner information processing. Informational structures of an agent become semantic information first in the case of highly intelligent agents.

Recently, empirical studies have revealed an unexpected richness of cognitive behaviors (perception, information processing, memory, decision making) in organisms as simple as bacteria. [18][19][32]. Single bacteria are too small, and sense only their immediate environment. They live too short to be able to memorize a significant amount of data. On the other hand bacterial colonies, swarms and films extends to a bigger space, have longer memory and exhibit an unanticipated complexity of behaviors that can undoubtedly be characterized as cognition [33][34][35]. Fascinating case are even simpler agents like viruses, on the border of the living [36][37]. Memory and learning are the key competences of living organisms [33].

Apart from bacteria and archaea [38] all other organisms without nervous system cognize (perceive their environment, process information, learn, memorize, communicate), such as e.g. slime mold, multinucleate or multicellular Amoebozoa, which has been used as natural computer to compute shortest paths. Even plants cognize, in spite of being typically thought of as living systems without cognitive capacities [39]. However, plants too have been found to possess memory (in their bodily structures that change as a result of past events), the ability to learn (plasticity, ability to adapt through morphodynamics), and the capacity to anticipate and direct their behavior accordingly. Plants are argued to possess rudimentary forms of knowledge, according to [40] p. 121, [41] p. 7 and [42] p. 61.

Consequently, in this article we take primitive cognition to be the totality of processes of self-generation/self-organization, self-regulation and self-maintenance that enables organisms to survive using information from the environment. The understanding of cognition as it appears in degrees of complexity in living nature can help us better understand the step between inanimate and animate matter from the first autocatalytic chemical reactions to the first autopoietic proto-cells.

## 4 LEARNING AS COMPUTATION IN NETWORKS OF AGENTS

Informational structures constituting the fabric of physical nature for an agent are networks of networks, which represent semantic relations between data. [17] Information is organized in layers, from quantum level to atomic, molecular, cellular/organismic, social, and so on. Computation/information processing, involve data structure exchanges within informational networks, represented by Carl Hewitt's actor model [43]. Different types of computation emerge at different levels of organization in nature as exchanges of informational structures between the nodes (computational agents). [11]

The research in computing nature/natural computing is characterized by bi-directional knowledge exchanges, through the interactions between computing and natural sciences. While natural sciences are adopting tools, methodologies and ideas of information processing, computing is broadening the notion of computation, taking information processing found in nature as computation. [2][44] Based on that, Denning argues that computing today is a natural science. [45] Computation found in nature is a physical process, where nature computes with physical bodies as objects. Physical laws govern processes of computation which appear on many different levels of organization.

With its layered computational architecture, natural computation provides a basis for a unified understanding of phenomena of embodied cognition, intelligence and learning (knowledge generation), including meta-learning. [30][46] Natural computation can be modelled as a *process of exchange of information in a network of informational agents* [43], that is entities capable of acting on their own behalf.

One sort of computation is found on the quantum-mechanical level where agents are elementary particles, and messages (information carriers) are exchanged by force carriers, while different types of computation can be found on other levels of organization in nature. In biology, information processing is going on in cells, tissues, organs, organisms and eco-systems, with corresponding agents and message types. In biological computing the message carriers are chunks of information such as molecules, while in social computing they are sentences while the computational nodes (agents) are be molecules, cells, organisms in biological computing or groups/societies in social computing. [18]

## 5 INFO-COMPUTATIONAL LEARNING BY MORPHOLOGICAL COMPUTATION

The notion of computation in this framework refers to the most general concept of *intrinsic computation,* that is a spontaneous computation processes in the nature, and which is used as a basis of specific kinds of *designed computation* found in computing machinery [47]. Intrinsic natural computation includes quantum computation [47][48], processes of self-organization, self-assembly, developmental processes, gene regulation networks, gene assembly, protein-protein interaction networks, biological transport networks, and similar. It is both analog (such as found in dynamic systems) and digital. The majority of info-computational processes are sub-symbolic and some of them are symbolic (like languages).

Within info-computational framework, computation on a given level of organization of information presents a realization/actualization of the laws that govern interactions between its constituent parts. On the basic level, computation is manifestation of causation in the physical substrate. In every next layer of organization a set of rules governing the system switch to the new emergent regime. It remains yet to be established how this process exactly goes on in nature, and how emergent properties occur. Research on natural computing is expected to uncover those mechanisms. In words of Rozenberg and Kari: "(O)ur task is nothing less than to discover a new, broader, notion of computation, and to understand the world around us *in terms of information processing*." [2] From the research in complex dynamical systems, biology, neuroscience, cognitive science, networks, concurrency and more, new insights essential for the info-computational universe may be expected.

Turing 1952 paper [49] may be considered as a predecessor of natural computing. It addressed the process of morphogenesis proposing a chemical model as the explanation of the development of biological patterns such as the spots and stripes on animal skin. Turing did not claim that physical system producing patterns actually performed computation. From the perspective of computing nature we can now argue that morphogenesis is a process of morphological computation. Informational structure (as representation of a physical structure) presents a *program* that governs computational process [50] which in its turn changes that original informational structure obeying/ implementing/ realizing physical laws.

*Morphology* is the central idea in our understanding of the connection between computation and information. Morphological/morphogenetic computing on that informational structure leads to new informational structures via processes of self-organization of information. Evolution itself is a process of morphological computation on a long-term scale. It is also possible to study morphogenesis of morphogenesis (Meta-morphogenesis) as done by Aaron Sloman in [51].

Leslie Valiant [52] studies evolution by ecorithms – learning algorithms that perform "probably approximately correct" PAC computation. Unlike classical paradigm of Turing computing, the results are not perfect, but *good enough* (for an agent).

## 6 LEARNING FROM RAW DATA AND UP – AGENCY FROM SYSTEM 1 TO SYSTEM 2

Cognition is a result of a processes of morphological computation on informational structures of a cognitive agent in the interaction with the physical world, with processes going on at both sub-symbolic and symbolic levels. This morphological computation establishes connections between an agent's body, its nervous (control) system and its environment. Through the embodied interaction with the informational structures of the environment, via sensory-motor coordination, information structures are induced (stimulated, produced) in the sensory data of a cognitive agent, thus establishing perception, categorization and learning. Those processes result in constant updates of memory and other structures that support behaviour, particularly *anticipation*. *Embodied* and corresponding *induced* (in the Sloman's sense of virtual machine) [53] informational structures are the basis of all cognitive activities, including consciousness and language as a means of maintenance of "reality" or the representation of the world.

From the simplest cognizing agents such as bacteria to the complex biological organisms with nervous systems and brains, the basic informational structures undergo transformations through morphological computation (developmental and evolutionary form generation), develop and evolve.

Living organisms as complex agents inherit bodily structures resulting from a long evolutionary development of species. Those structures are embodied memory of the evolutionary past. They present the means for agents to interact with the world, get new information that induces memories, learn new patterns of behaviour and learn/construct knowledge. By Hebbian learning, world shapes human's (or an animal's) informational structures., Neural networks that "self-organize stable pattern recognition codes in real-time in response to arbitrary sequences of input patterns" are illustrative example. [54]

If we say that for something to be information there must exist an agent from whose perspective this structure is established, and we argue that the fabric of the world is informational, the question can be asked: *who/what is the agent*? An agent (an entity capable of acting on its own behalf) can be seen as interacting with the points of inhomogeneity (data), establishing the connections between those data and the data that constitute the agent itself (a particle, a system). There are myriads of agents for which information of the world makes differences – from elementary particles to molecules, cells, organisms, societies… - all of them interact and exchange information on different levels of scale and this information dynamics is natural computation.

On the fundamental level of quantum mechanical substrate, information processes represent actions of laws of physics. Physicists are already working on reformulating physics in terms of information [53]. This development can be related to the Wheeler's idea "it from bit". [55] and von Weizsäcker's ur-alternatives [56].

## 11 CONCLUSIONS AND FUTURE WORK

Contemporary Deep-Learning-Centered AI is developing from the present state System 1 coverage towards the System 2, with agency, causality, attention and consciousness as mechanisms of learning and meta-learning (learning to learn). In this process like in the past, deep learning is searching inspiration in nature, assimilating ideas from neuroscience, cognitive science, biology, and more. This approach to understanding, via decomposition and construction is close to other computational models of nature in that it seeks testable and applicable models, based on data and information processing.

At the same time, Computing nature approach models nature as consisting of physical structures that form levels of organization, on which computation processes develop. It has been argued that on the lower levels of organization finite automata or Turing machines might be an adequate model, while on the level of the whole-brain non-Turing computation is necessary, Ehresmann [57] and Ghosh et al. [58]

Within info-computational framework, cognition is synonymous with the process of life, which enables learning from life characteristics to cognitive properties within evolutionary process. As mentioned before, evolution is learning process where nature tests varieties of possibilities. Following Maturana and Varela [21], we understand the entire living word as possessing cognition of various degrees of complexity. In that sense bacteria possess rudimentary cognition expressed in quorum sensing and other collective phenomena based on information communication and information processing. Brain of a complex organism consists of neurons that are networked, communicational and computational units. Signalling and information processing modes of a brain are much more complex and consist of more info-computational layers than bacterial colony. Knowledge of the world for an agent is an informational structure that is established as a result of as well the interactions of the agent with the environment (System 1) as the information processes in agents own intrinsic structures – reasoning, anticipation, etc. (System 2).

For the future, work remains to be done on the connections between the low level and the high level cognitive processes. It is also important to find relations between cognition and consciousness as a mechanism helping to reduce number of variables that are manipulated by an agent (an organism) for the purpose of reasoning, decision-making, planning and acting in the world.

The goals of AI different from the goals of the computing nature framework. AI builds solutions for practical problems and in that it focus on (typically highest possible level of) intelligence (not yet emotional nor embodied intelligence at this stage of the development), while computing nature framework seeks to provide computational models of all kinds of natural systems, including living organisms and their evolution and development, with not only intelligence but also full scale of cognition with emotion and behaviours that are not always goal-oriented in the sense of AI. The priority of info-computational naturalism is understanding and connecting knowledge about nature, while for AI the priority is practical problem solving. Nevertheless, paths of the two are meeting in many cases and mutual exchange of ideas promises benefits for both.

## REFERENCES

[1] Y. Lecun, Y. Bengio, and G. Hinton, "Deep learning," Nature. 2015.
[2] G. Rozenberg and L. Kari, "The many facets of natural computing," Commun. ACM, vol. 51, pp. 72–83, 2008.
[3] G. Dodig-Crnkovic, "Nature as a network of morphological infocomputational processes for cognitive agents," Eur. Phys. J. Spec. Top., 2017.
[4] G. Dodig-Crnkovic, "Cognition as Embodied Morphological Computation," in *Philosophy and Theory of Artificial Intelligence*, vol. 44, Studies in Applied Philosophy, Epistemology and Rational Ethics, 2018, pp. 19–23.
[5] Y. Bengio, "From System 1 Deep Learning to System 2 Deep Learning (NeurIPS 2019)," 2019. [Online]. Available: https://www.youtube.com/watch?v=T3sxeTgT4qc.
[6] Y. Bengio, "Scaling up deep learning," 2014.
[7] Y. Bengio, "The Consciousness Prior," *arXiv:1709.08568v2*, 2019.
[8] D. Kahneman, *Thinking, Fast and Slow*. New York: Farrar, Straus and Giroux, 2011.
[9] A. Clark, *Microcognition: Philosophy, Cognitive Science, and Parallel Distributed Processing*. Cambridge, MA: MIT Press, 1989.
[10] B. Scellier and Y. Bengio, "Towards a Biologically Plausible Backprop," *Arxiv*, 2016.
[11] G. Dodig-Crnkovic, "Information, Computation, Cognition. Agency-Based Hierarchies of Levels," in *Fundamental Issues of Artificial Intelligence. Synthese Library, (Studies in Epistemology, Logic, Methodology, and Philosophy of Science), vol 376.*, V. Müller, Ed. Springer International Publishing, 2016, pp. 141–159.
[12] S. Kauffman, *Origins of Order: Self-Organization and Selection in Evolution*. Oxford University Press, 1993.
[13] T. Deacon, *Incomplete Nature. How Mind Emerged from Matter*. New York. London: W. W. Norton & Company, 2011.
[14] H. Maturana, "Biology of Cognition," Defense Technical Information Center, Illinois, 1970.
[15] H. Maturana and F. Varela, *The Tree of Knowledge*. Shambala, 1992.
[16] J. Stewart, "Cognition = life: Implications for higher-level cognition," *Behav. Process.*, vol. 35, pp. 311-326., 1996.
[17] G. Dodig-Crnkovic and R. Giovagnoli, *COMPUTING NATURE*, vol. 7. Springer, 2013.
[18] G. Dodig-Crnkovic, "Physical computation as dynamics of form that glues everything together," *Inf.*, 2012.
[19] E. Ben-Jacob, "Bacterial Complexity: More Is Different on All

Levels," in *Systems Biology- The Challenge of Complexity*, S. Nakanishi, R. Kageyama, and D. Watanabe, Eds. Tokyo Berlin Heidelberg New York: Springer, 2009, pp. 25–35.
[20] E. Ben-Jacob, "Learning from Bacteria about Natural Information Processing," *Ann. N. Y. Acad. Sci.*, vol. 1178, pp. 78–90, 2009.
[21] H. Maturana and F. Varela, *Autopoiesis and cognition: the realization of the living*. Dordrecht Holland: D. Reidel Pub. Co., 1980.
[22] M. van Duijn, F. Keijzer, and D. Franken, "Principles of Minimal Cognition: Casting Cognition as Sensorimotor Coordination," *Adapt. Behav.*, vol. 14, no. 2, pp. 157–170, 2006.
[23] P. Århem and H. Liljenström, "On the coevolution of cognition and consciousness," *J. Theor. Biol.*, 1997.
[24] H. Liljenström and P. Århem, *Consciousness Transitions: Phylogenetic, Ontogenetic and Physiological Aspects*. Amsterdam: Elsevier, 2011.
[25] G. Dodig-Crnkovic and von H. Rickard, "Reality Construction in Cognitive Agents through Processes of Info-Computation," in *Representation and Reality in Humans, Other Living Organisms and Intelligent Machines*, G. Dodig-Crnkovic and R. Giovagnoli, Eds. Basel: Springer International Publishing, 2017, pp. 211–235.
[26] H. Putnam, *Mathematics, Matter and Method*. Cambridge: Cambridge University Press, 1975.
[27] G. Dodig-Crnkovic and M. Schroeder, "Contemporary Natural Philosophy and Philosophies," *Philosophies*, vol. 3, no. 4, p. 42, Nov. 2018.
[28] G. Dodig-Crnkovic and M. Schroeder, *Contemporary Natural Philosophy and Philosophies - Part 1*. Basel: MDPI AG, 2019.
[29] G. Dodig-Crnkovic, *Investigations into Information Semantics and Ethics of Computing*. Västerås, Sweden: Mälardalen University Press, 2006.
[30] G. Dodig-Crnkovic and V. Müller, "A Dialogue Concerning Two World Systems: Info-Computational vs. Mechanistic," World Scientific Pub Co Inc, Singapore, 2009.
[31] G. Dodig-Crnkovic, "The info-computational nature of morphological computing," in *Studies in Applied Philosophy, Epistemology and Rational Ethics*, 2013.
[32] W.-L. Ng and B. L. Bassler, "Bacterial quorum-sensing network architectures," *Annu. Rev. Genet.*, vol. 43, pp. 197–222, 2009.
[33] G. Witzany, "Memory and Learning as Key Competences of Living Organisms," 2018.
[34] G. Witzany, "Introduction: Key Levels of Biocommunication of Bacteria," 2011.
[35] D. Dennett, *From Bacteria to Bach and Back: The Evolution of Minds*. W. W. Norton & Company, 2017.
[36] G. Witzany, *Viruses: Essential agents of life*. 2012.
[37] L. P. Villarreal and G. Witzany, "Viruses are essential agents within the roots and stem of the tree of life," *J. Theor. Biol.*, 2010.
[38] G. Witzany, *Biocommunication of archaea*. 2017.
[39] G. Witzany, "Bio-communication of Plants," *Nat. Preced.*, 2007.
[40] R. S. Pombo, O., Torres J.M., Symons J., Ed., *Special Sciences and the Unity of Science*, Logic, Epi. Berlin Heidelberg: Springer, 2012.
[41] R. Rosen, *Anticipatory Systems*. New York: Pergamon Press, 1985.
[42] K. Popper, *All Life Is Problem Solving*. London: Routledge, 1999.
[43] C. Hewitt, "What is computation? Actor Model versus Turing's Model," in *A Computable Universe, Understanding Computation & Exploring Nature As Computation*, H. Zenil, Ed. World Scientific Publishing Company/Imperial College Press, 2012.
[44] G. Rozenberg, T. Bäck, and J. N. Kok, Eds., *Handbook of Natural Computing*. Berlin Heidelberg: Springer, 2012.
[45] P. Denning, "Computing is a natural science," *Commun. ACM*, vol. 50, no. 7, pp. 13–18, 2007.
[46] Y. Wang, "On Abstract Intelligence: Toward a Unifying Theory of Natural, Artificial, Machinable, and Computational Intelligence," *Int. J. Softw. Sci. Comput. Intell.*, vol. 1, no. 1, pp. 1–17, 2009.
[47] S. Crutchfield, James P.; Ditto, William L.; Sinha, "Introduction to Focus Issue: Intrinsic and Designed Computation: Information Processing in Dynamical Systems-Beyond the Digital Hegemony," *Chaos*, vol. 20, no. 3, pp. 037101-037101–6, 2010.
[48] J. P. Crutchfield and K. Wiesner, "Intrinsic Quantum Computation," *Phys. Lett. A*, vol. 374, no. 4, pp. 375–380, 2008.
[49] A. M. Turing, "The Chemical Basis of Morphogenesis," *Philos. Trans. R. Soc. London*, vol. 237, no. 641, pp. 37–72, 1952.
[50] G. Kampis, *Self-modifying systems in biology and cognitive science: a new framework for dynamics, information, and complexity*. Amsterdam: Pergamon Press, 1991.
[51] A. Sloman, "Meta-Morphogenesis: Evolution and Development of Information-Processing Machinery p. 849.," in *Alan Turing: His Work and Impact*, S. B. Cooper and J. van Leeuwen, Eds. Amsterdam: Elsevier, 2013.
[52] L. Valiant, *Probably Approximately Correct: Nature's Algorithms for Learning and Prospering in a Complex World*. New York: Basic Books, 2013.
[53] A. Sloman and R. Chrisley, "Virtual machines and consciousness," *J. Conscious. Stud.*, vol. 10, no. 4–5, pp. 113–172, 2003.
[54] G. A. Grossberg and S. Carpenter, "ART 2: self-organization of stable category recognition codes for analog input patterns," *Appl. Opt.*, vol. 26, no. 23, pp. 4919–4930, 1987.
[55] J. A. Wheeler, "Information, physics, quantum: The search for links," in *Complexity, Entropy, and the Physics of Information*, W. Zurek, Ed. Redwood City: Addison-Wesley, 1990.
[56] C. F. Weizcsäcker, "The Unity of Nature," in *Physical Sciences and History of Physics. Boston Studies in the Philosophy of Science. vol. 82.*, C. R.S. and M. W. Wartofsky, Eds. Springer, Dordrecht, 1984.
[57] A. C. Ehresmann, "MENS, an Info-Computational Model for (Neuro-)cognitive Systems Capable of Creativity," *Entropy*, vol. 14, pp. 1703-1716., 2012.
[58] S. Ghosh, K. Aswani, S. Singh, S. Sahu, D. Fujita, and A. Bandyopadhyay, "Design and Construction of a Brain-Like Computer: A New Class of Frequency-Fractal Computing Using Wireless Communication in a Supramolecular Organic, Inorganic System," *Information*, vol. 5, no. 1, pp. 28–100, Jan. 2014.